\documentclass[conference]{IEEEtran}
\IEEEoverridecommandlockouts
\usepackage{cite}
\usepackage{url}
\usepackage{amsmath,amssymb,amsfonts}
\usepackage{algorithmic}
\usepackage{algorithm}
\usepackage{multirow}
\usepackage{graphicx}
\usepackage{textcomp}
\usepackage{xcolor}
\usepackage{hyperref} 
\usepackage{setspace}
\usepackage{caption}
\usepackage{subcaption}

\renewcommand{\IEEEbibitemsep}{0pt plus 0.5pt}
\makeatletter
\IEEEtriggercmd{\reset@font\normalfont\fontsize{7.9pt}{8.40pt}\selectfont}
\makeatother
\IEEEtriggeratref{1}
\def\BibTeX{{\rm B\kern-.05em{\sc i\kern-.025em b}\kern-.08em
    T\kern-.1667em\lower.7ex\hbox{E}\kern-.125emX}}
    
\makeatletter
\def\@IEEEpubidpullup{8\baselineskip}
\makeatother
\begin{document}
\IEEEoverridecommandlockouts
\IEEEpubid{
\parbox{\columnwidth}{\vspace{-4\baselineskip}Permission to make digital or hard copies of all or part of this work for personal or classroom use is granted without fee provided that copies are not made or distributed for profit or commercial advantage and that copies bear this notice and the full citation on the first page. Copyrights for components of this work owned by others than ACM must be honored. Abstracting with credit is permitted. To copy otherwise, or republish, to post on servers or to redistribute to lists, requires prior specific permission and/or a fee. Request permissions from \href{mailto:permissions@acm.org}{permissions@acm.org}.\hfill\vspace{-0.8\baselineskip}\\
\begin{spacing}{1.2}
\small\textit{ASONAM '19}, August 27-30, 2019, Vancouver, Canada \\ 
\copyright\space2019 Association for Computing Machinery. \\
ACM ISBN 978-1-4503-6868-1/19/08\$15.00 \\
\url{http://dx.doi.org/10.1145/3341161.3342936} 
\end{spacing}
\hfill}
\hspace{0.9\columnsep}\makebox[\columnwidth]{\hfill}}
\IEEEpubidadjcol

\title{Low-supervision urgency detection and transfer in short crisis messages
}

\author{\IEEEauthorblockN{Mayank Kejriwal}
\IEEEauthorblockA{\textit{Information Sciences Institute} \\
\textit{University of Southern California}\\
Marina del Rey, CA \\
kejriwal@isi.edu}
\and
\IEEEauthorblockN{Peilin Zhou}
\IEEEauthorblockA{\textit{Information Sciences Institute} \\
\textit{University of Southern California}\\
Marina del Rey, CA \\
zpeilin@isi.edu}
}

\maketitle

\begin{abstract}
Humanitarian disasters have been on the rise in recent years due to the effects of climate change and socio-political situations such as the refugee crisis. Technology can be used to best mobilize resources such as food and water in the event of a natural disaster, by semi-automatically flagging tweets and short messages as indicating an urgent need. The problem is challenging not just because of the sparseness of data in the immediate aftermath of a disaster, but because of the varying characteristics of disasters in developing countries (making it difficult to train just one system) and the noise and quirks in social media. In this paper, we present a robust, low-supervision social media urgency system that adapts to arbitrary crises by leveraging both labeled and unlabeled data in an ensemble setting. The system is also able to adapt to new crises where an unlabeled background corpus may not be available yet by utilizing a simple and effective transfer learning methodology.
Experimentally, our transfer learning and low-supervision approaches are found to outperform viable baselines with high significance on myriad disaster datasets.  
\end{abstract}


\section{Introduction}


\begin{table*}
\centering
\caption{Urgent and non-urgent examples from three real-world datasets that we describe further in Section \ref{experiments}.}
\begin{tabular}{|p{0.65in}|p{2.9in}| p{3.0in} |} \hline
Dataset & Urgent Sentences & Non-urgent Sentences \\ \hline 
Nepal & Anyone who speaks about Balochistan in provinces other than Punjab either ends up dead or missing & Today's earthquake data for Nepal \\ \hline 
 & EMERGENCY: 4 locals trapped in this rubble INSIDE PALTANGHAR & Wow. ndtv just showed the same Philippines earthquake picture and said it's from Kathmandu on TV.  \\ \hline 
Macedonia & Some people are trapped in the marketplace need help. & the streets are filled with fecal and water no water  \\ \hline 
 & We re trapped at the national commissioner s house the first floor s loaded with the kids have begun scared. & I'm about to walk with bicite but the rain that fell before s been blocking the roads that the channels are from the time of the rock.  \\ \hline 
Kerala & 8 people no food survivin on dry cornflakes for the last 3 days east kadungalloor two families. & I'm from kerala and the situation here is very very bad, thousands have lost.  \\ \hline 
 & At least 324 people have been killed in flooding and landslides in the indian state of while more than 200000 & there has been floods in kerala india, more than 70 have lost their lives may  "Make it easy for all".  \\ \hline 

\end{tabular}
\end{table*}


The United Nations Office for the Coordination of Human Affairs (OCHA) reported\footnote{\url{https://www.unocha.org/sites/unocha/files/WHDT2018_web_final_spread.pdf}} that in 2018, more than 141 million people were in need of humanitarian assistance, with over 9 billion dollars of unmet requirements. Using technology to address this shortfall by assisting aid agencies and first responders mobilize and send resources where they are needed the most is an important problem with the potential for widespread long-lasting social impact \cite{policyforum}, \cite{earthquake1}. 

To achieve this goal, the problem of semi-automatic \emph{urgency detection} needs to be solved, especially on short message streams like social media that support real-time news feeds and micro-updates from citizens on the ground. Put intuitively, the urgency detection problem can be framed in terms of probabilistic binary classification, a common machine learning paradigm involving other related tasks like sentiment analysis \cite{sentimentanalysis}. Although urgency detection has some similarity with sentiment analysis, the core problem is different, since the goal is to flag messages that \emph{express urgency}, which is almost always a negative or panic-ridden emotion. However, it can be difficult to distinguish urgency-related tweets from just negative tweets. We provide an illustrative set of real-world examples\footnote{A description of the datasets will be provided in Section \ref{data}.} in Table I.

In this paper, we present practical approaches for crisis-specific minimally supervised urgency detection on short message streams such as Twitter. The presented approaches cover two scenarios that often emerge in the real world. In the first scenario, a small amount (a few hundred messages) of training data labeled as urgent or non-urgent is available, along with a copious `unlabeled' background corpus. In the second scenario, similar data is available for a `source' domain but not for the target domain (expressing a `new crisis') for which the urgency detection needs to be deployed. In other words, as messages are streaming in for this new domain, investigators label a few samples, but cannot rely on the availability of a background corpus since urgency needs to be tagged in real time before the crisis has fully subsided. To accomplish this challenging goal, our approach relies on a simple and robust transfer learning methodology \cite{TLsurvey}. Experimental results on three real-world datasets and several performance metrics validate our methods. To the best of our knowledge, this is the first such paper investigating the problem of urgency detection in social media, both algorithmically and empirically,  for arbitrary disasters in low-supervision and transfer learning settings. 

The rest of this paper is structured as follows. Section \ref{relatedwork} describes some related work, Section \ref{RQs} specifies our two research questions, and Section \ref{approach} describes our approaches in support of answering those questions. Section \ref{experiments} covers the experiments, and Section \ref{conclusion} concludes the paper.

%
%

 
 \section{Related Work}\label{relatedwork}

\emph{Crisis informatics} is emerging as an important field for both data scientists and policy analysts. A good introduction to the field was provided in a recent Science policy forum article~\cite{policyforum}. The field draws on interdisciplinary strands of research, especially with respect to collecting, processing and analyzing real-world data. Particularly, social media platforms like Twitter have emerged as important channels (`social sensors'~\cite{earthquake1}) for \emph{situational awareness} in support of crisis informatics. Although situational awareness is a broad notion extending beyond crisis informatics (e.g., military situational awareness), urgency detection is a special kind of situational awareness that tends to arise mainly in the crisis domain. A direct application is to help first responders and aid agencies assess needs in crisis-stricken areas and mobilize resources effectively (i.e. where needs are most urgent).
NLP methods have been widely used in extracting situational awareness from Twitter e.g., see the work by Verma et al.~\cite{verma}. Another important line of work is in analyzing events other than natural disasters (such as mass convergence and disruption events), but still relevant to crisis informatics. For example Stabird et al. presented a collaborative filtering system for identifying on-the-ground `Twitterers' during mass disruptions~\cite{starbird}. Similar techniques could be employed to supplement the work in this paper. 

More generally, projects like CrisisLex, Crisis Computing\footnote{\url{https://crisiscomputing.qcri.org/}} and EPIC (Empowering the Public with Information in Crisis) have emerged as major efforts in the crisis informatics space due to two reasons: first, the abundance and fine granularity of social media data implies that mining such data during crises can lead to robust, real-time responses; second, the recognition that any technology that is thus developed must also address the inherent challenges (including problems of noise, scale and irrelevance) in working with such datasets. CrisisLex provides a repository of crisis-related social media data and tools, including collections of crisis data and lexicons of crisis terms~\cite{olteanu2014crisislex}. It also includes tools to help users create their own collections and lexicons. In contrast, Project EPIC,
~launched in 2009 and supported by a US National Science Foundation grant, is a multi-disciplinary effort involving several universities and languages with the goal of utilizing behavioral and technical knowledge of computer mediated communication for better crisis study and emergency response. Since its founding, Project EPIC has led to several advances in the crisis informatics space; see for example~\cite{epic1,epic2,epic3,epic4,epic5}. The work presented in this article is intended to be compatible with these efforts. 

Other lines of work relevant to this paper involve minimally supervised machine learning, representation learning and transfer learning. Concerning minimally supervised machine learning (ML), in general, ML techniques where there are few, and in the case of zero-shot learning~\cite{zs1,zs2}, no observed instances for a label has been a popular research agenda for many years~\cite{minsup,weaksup}. In addition to weak supervision approaches~\cite{weaksup}, both semi-supervised and active learning have also been studied in great depth, with surveys provided by~\cite{semisup,activelearning}. However, to the best of our knowledge, a successful systems-level conjunction of various minimally supervised ML techniques has not been achieved for the task of short-text urgency detection. Such as empirical assessment is an important goal of this paper. 

Due to the current renaissance of neural networks~\cite{embedding:old}, \emph{embedding} and \emph{representation learning} methods have become more popular due to the advent of fast and effective models like skip-gram. Recent work has used such embeddings in numerous NLP and graph-theoretic applications~\cite{nlpfromscratch}, including information extraction~\cite{kejriwalIE}, named entity recognition~\cite{ner} and entity linking~\cite{entitylinking}. The most well-known example is word2vec (for words)~\cite{word2vec}, followed by similar models like paragraph2vec (for multi-word text) and fasttext~\cite{docvec,fasttext}, the last two being most relevant for the work in this paper. For a recent evaluation study on representation learning for text, including potential problems, we refer the reader to~\cite{RLsurvey}.  Finally, transfer learning is a central agenda in this paper; an excellent survey of dominant techniques may be found in \cite{TLsurvey}. More recent work on domain adaptation may be found in \cite{suggested6}, with the work in \cite{suggested5} applied specifically to the disaster response problem. Pedrood and Purohit \cite{suggested5} also applied transfer learning to the problem of mining help intent on Twitter. Other relevant work in crisis informatics, both in terms of defining `actionable information' problems like urgency and need mining, as well as providing multimodal Twitter datasets from natural disasters, may be found in \cite{suggested1}, \cite{suggested2} and \cite{suggested3}. An alternate way of looking at the problem is as an `event detection' problem e.g., in \cite{suggested7} Zheng et al. study semi-supervised event-related tweet identification which also tries to identify the urgent tweets related to earthquakes and floods. These works are complementary to the minimally supervised, low-resource setting in this paper.

\section{Research Questions}\label{RQs}
We briefly enumerate below the research questions under consideration in this paper. While the first question captures the classical low-supervision setting, the second question introduces an element of transfer learning. 

\begin{enumerate}
    \item {\bf Low-supervision Training for Urgency Detection:} How do we build an urgency detection system for a specific crisis when given as training input both a small number of manually labeled tweets, and a large number of unlabeled tweets (background corpus), for that crisis?
    \item {\bf Low-supervision Transfer Learning for Urgency Detection:} How do we build an urgency detection system for a specific crisis when given as training input a small number of manually labeled tweets for that crisis, as well as `auxiliary' training input of (a small number of) manually labeled tweets and unlabeled background tweets from a \emph{different} crisis?
\end{enumerate}

Unlike the first scenario, the second scenario applies to a very short period (hours, or even minutes) after the crisis has struck; this is why a background corpus is not available (yet) for that crisis. Instead, only a few manually labeled messages that have been acquired till that point are available.

\section{Approach}\label{approach}



\subsection{Low-supervision urgency detection}
The approach for addressing the first research question is schematized in Figure 1. The first step in the workflow involves data preprocessing of the corpus. We follow a standard set of preprocessing steps. First, we apply a tokenizer to split the sentences into lists of words and delete words with special prefixes (including @ and RT, which are particularly prevalent in Twitter), and special suffixes. We also remove non-alphanumeric characters and convert the entire sentence to lowercase.
Next, similar to traditional machine learning pipelines, we extract a set of manual features for expressing prior human knowledge about urgency detection. Our manual features are thus called because they are primarily keyword-based and binary, with keywords selected based on data exploration and domain knowledge. We consider ten such keywords, namely \emph{hit, help, kill, injure, strand, miss, urgent, die, need, food}. If any of these keywords are present\footnote{Possibly as stems, for example, the word `helping' would trigger the `help' keyword feature, which would be consequently set to 1.}, the corresponding feature is set to 1. Note that these keywords are associated with situations that are generally urgent, like people who have been attacked or affected by a crisis and need urgent help, but some are noisier than others\footnote{For example `help' could be associated with a more trivial situation like someone needing help with their dog.}. Additionally, we also utilize an eleventh feature that checks to see if any numeric digits are present in the dataset. The rationale behind this feature is that, in more urgent tweets, numbers are often present e.g., `15 climbers are currently trapped on Everest due to the avalanche'.  



In the experimental section, we show that the manual features are not adequate for addressing low-supervision urgency detection. Besides, it is prudent to utilize the large number of unlabeled tweets (background corpus) if it serves a useful purpose in improving performance. To that end, we train a skip-gram based word embedding model based on the `bag of tricks' model released by researchers from Facebook in a package called \emph{fastText} \cite{fasttext}. The reason behind using fastText, as opposed to alternate word embedding models like GloVe and word2vec \cite{word2vec}, is several-fold. First, fastText is very fast and easy to execute, and is well-maintained. Second, preliminary analyses showed that it does quite well on social media tasks and because of the bag of tricks methodology (that uses character and sub-word embeddings to gracefully deal with OOVs\footnote{Out of Vocabulary words.} and misspellings), it is able to generalize much better. Finally, fastText's APIs include a way to get sentence embeddings directly after training the word embedding model. By training fastText on the background corpus, we are able to train a robust embedding model. In both the training and test phase, we use this model to get feature vectors for our messages besides the 11-dimensional manual feature vector described earlier.   

However, given that the background corpus might not be as extensive or representative as a `general' corpus like Wikipedia, we try to smooth the feature space by also using a pre-trained embedding model trained over the English Wikipedia corpus and publicly available\footnote{\url{https://fasttext.cc/docs/en/pretrained-vectors.html}}. The vectors obtained from this model have 300 dimensions and were trained using skip gram with default parameters.

As Figure 1 illustrates, we use all of these feature sets to build an ensemble by combining local embedding features, manual features and Wikipedia pre-trained word embedding features. The final score of the ensemble model is achieved by weighting the scores of the three Linear Regression models (one for each feature-set), with weights adding to 1. The weights are set using a held-out validation set.

When the urgency of a new `test' message needs to be determined, we preprocess the message, extract all three feature-sets\footnote{In the case of the two trained embedding models, by getting the respective sentence embeddings for the test message}, and get the weighted score from the three regression models. If the score falls above a pre-determined threshold (again, determined through validation), then the message is flagged as urgent, otherwise it is not.




\begin{figure}[htbp]
\centering
\begin{minipage}[]{0.5\textwidth}
\includegraphics[width=1.0\textwidth,angle=0]{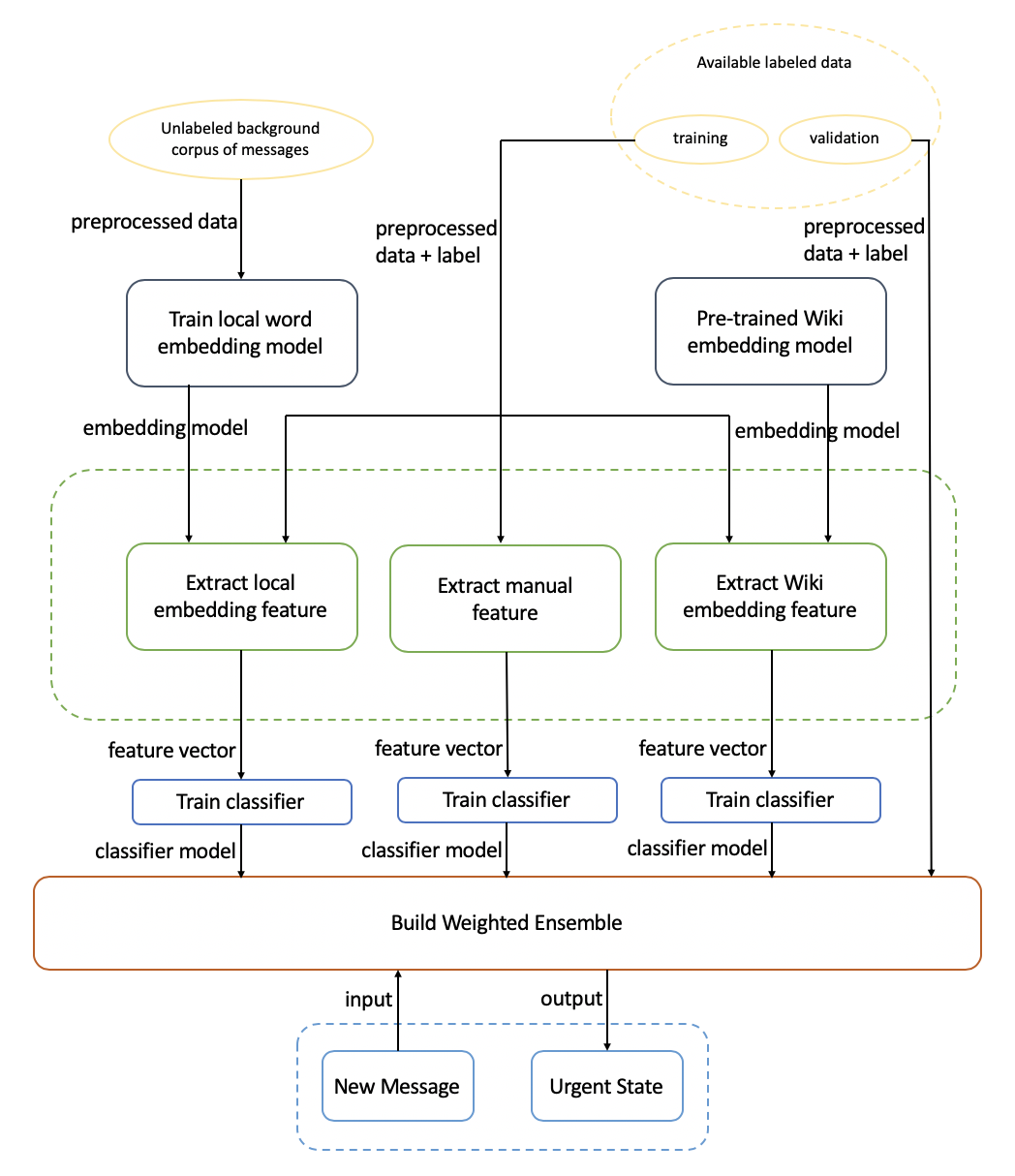}
\end{minipage}
\caption{Training for Urgency Detection.}
\end{figure}

\subsection{Urgency detection using transfer learning}
In this section, we describe our approach for `urgency detection transfer' whereby a \emph{source} dataset is given (similar to RQ1, where both an unlabeled background corpus, as well as a small manually labeled training set, are available) along with a \emph{target} dataset (only a small manually labeled training set and no background corpus), representing the crisis under investigation. Our approach for urgency transfer is captured in Algorithm 1. Many of the steps are similar to those for RQ1, including preprocessing, but there are some important differences. For example, while the Wiki embedding model remains the same as earlier, the manual features are obviously extracted over the target domain (since they do not require a background corpus) and importantly, the `local' embedding model is now trained over the source domain corpus, since there is no target domain unlabeled background corpus available.

To `sync' the source and target domains, we consider a simple, but empirically effective, approach. Rather than use just the labeled target domain data for training the three linear regression models, we combine the labeled training data from both the source and target domains, but the target training data is up-sampled to allow its properties to emerge more concretely in the training. The up-sampling margin is a parameter in Algorithm 1; in practice, a factor of 6 (meaning the target labeled dataset is up-sampled by 6x) has been found to work well. To maximize training dataset utility, we do not use a validation set for classifier weight optimization, but consider the average of all three classifiers as the final score.

\begin{algorithm}[htb] 
\caption{ Transfer Learning for Urgency Detection.} 
\label{alg:Framwork} 
\begin{algorithmic}
\STATE \textbf{Input :} \begin{itemize}
\item Labeled dataset in target domain: $D_t$\\
\item Labeled dataset in source domain: $D_{sl}$ \\
\item Unlabeled corpus in source domain: $D_{su}$\\
\item Pre-trained Wikipedia Embedding Model: $W_w$ \\
\item Up-sampling parameter: $u$
\end{itemize}
\STATE \textbf{Output :} \begin{itemize}
\item Classifier for Urgency Detection: $\mathcal{C}$
\end{itemize}
\STATE \textbf{Method :}
\begin{enumerate}
\STATE Train word embedding $W_s$ on text in $D_{su} \cup D_{sl}$ ;
\STATE Up-sample $D_t$ by factor $u$ and `mix' with $D_{sl}$ to get expanded training set, $D_{train}: D_{tu} \cup D_sl$
\STATE Extract manual feature set $F_m$, source embedding feature set $F_s$ (using $W_s$), and Wiki feature set $F_w$ (using $W_w$) from each message in $D_{train}$;
\STATE Train linear regression models $C_s$, $C_m$ and $C_w$ on $F_s$, $F_m$ and $F_w$ resp. to get classifier;
\STATE Return final classifier model $\mathcal{C}: avg\_score(C_s, C_m,C_w)$;
\end{enumerate}
\end{algorithmic}
\end{algorithm}

\section{Experiments}\label{experiments}

\subsection{Data}\label{data}
\begin{table*}
\centering
\caption{Details on datasets used for experiments.}
\begin{tabular}{|p{0.7in}|p{1.2in}| p{1.0in}| p{0.6in}| p{0.8in}|p{1.5in}|} \hline
Dataset  &Unlabeled / Labeled Messages & Urgent / Non-urgent Messages & Unique Tokens & Avg. Tokens / Message & Time Range \\ \hline
 Nepal & 6,063/400 & 201/199 & 1,641 & 14 & 04/05/2015-05/06/2015\\ \hline
 Macedonia & 0/205 & 92/113 & 129 & 18 & 09/18/2018-09/21/2018\\ \hline
 Kerala & 92,046/400 & 125/275 & 19,393 & 15 & 08/17/2018-08/22/2018\\ \hline
\end{tabular}\label{table:datasets}
\end{table*}
For evaluating the approaches laid out in Section \ref{approach}, we consider three real-world datasets described in Table \ref{table:datasets}.

Two of the datasets (Nepal and Macedonia) were made available to us through the DARPA LORELEI program, under which this project is funded. The Nepal dataset comprises a collection of tweets collected in the aftermath of the 2015 Nepal earthquake (also called the Gorkha earthquake), while Macedonia was not an actual disaster but a realistic live-action simulation (of a disaster) conduced in Macedonia towards the end of 2018. Macedonia does not have much noise and is `information-dense', but small. As such, it provides a good test of the transfer learning abilities of the approach presented. Kerala describes tweets in the aftermath of the Kerala floods in South India in 2018, and is the largest dataset, with many relevant and irrelevant tweets. 

Originally, all the raw messages for the datasets described in Table \ref{table:datasets} were unlabeled, in that their urgency status was unknown. Since the Macedonia dataset only contains 205 messages, and is a small but information-dense dataset, we labeled all messages in Macedonia as urgent or non-urgent (hence, there are no unlabeled messages in Macedonia per Table \ref{table:datasets}).  For the two other Twitter-based datasets, we used active learning to compose a labeled set that would contain challenging examples. The basic process was to do data preprocessing as described in Section \ref{approach}, followed by training the local fastText-based word embedding model on all messages in the corpus. Next, we randomly labeled 50 urgent and non-urgent tweets and fed them into a classifier. The classifier was applied on the rest of the unlabeled data to obtain `ambiguous' examples (where the classifier's probability of the positive label was closest to 50\%). We labeled another 100 samples this way, and continued to re-train and apply the classifier for two more iterations till we obtained a total of 400 labeled points. 
Note that the final labeled dataset may not be balanced in terms of urgent and non-urgent messages. Table \ref{table:datasets} shows that Nepal is roughly balanced, while Kerala is imbalanced. We used stratified sampling therefore to split the labeled pool into a training and testing dataset for evaluating the two research questions. We used 90\% for training and 10\% for testing.

\subsection{Metrics}

We consider four standard metrics, namely \emph{Accuracy, Precision, Recall and F-Measure}. Accuracy is simply the ratio of correctly labeled messages to the size of test set, precision is the ratio of the true positives to the sum of true positives and false positives, recall is the ratio of true positives to the sum of true positives and false negatives, and finally, F-Measure is the harmonic mean of precision and recall and captures their trade-off. 

\subsection{Methodology}

\subsubsection{Protocol}
Concerning RQ1, for datasets, we use \emph{Nepal} and \emph{Kerala} since \emph{Macedonia} does not have a large unlabeled corpus available, which is an assumption made per RQ1. Recall that we used stratified random sampling to split the labeled data for each dataset into training (90\%) and test (10\%) sets. Of the 90\% training set, a further split was done, with 90\% kept for `training' and 10\% for setting optimal weights for the 3 linear classifiers\footnote{The hyperparameters of the linear regression itself were optimized through 5-fold cross-validation on this `inner' (i.e. 90\% of the original 90\% training set) training set.} trained in Section IV. To account for the effects of randomness, each experiment was conducted across ten trials, with averages reported on all four metrics described previously for all baselines described below and our approach. Among the different machine learning classifiers in the sklearn package tested, the linear regression was found to work well and used as the classifier of choice where applicable.

\subsubsection{Baselines for Low-supervision Training for Urgency Detection}
We use six baselines to evaluate the approach for RQ1 described in Section \ref{approach}. Note that statistical significance is tested using the one-sided Student's paired t-test by comparing the best system (on each metric) against the \emph{Local} baseline, which is a reasonable choice since in a high-supervision (or even normal-supervision) setting, this baseline has been found to perform quite well. Significance at the 90\% level is indicated with a *, at the 95\% level with a **, and at the 99\% level with a ***. 

\begin{table*}
\centering
\caption{Description For Each Baseline On Research Question 1.}
\begin{tabular}{|p{2.5in}|p{4.5in}| } \hline
Baseline & Description \\ \hline 
Local Embedding (Local) & The features for a single linear classifier are sentence embeddings (with each pre-processed message treated as a `sentence') trained using the 5-gram skip gram-based fastText model with vector dimensionality set to 20 \\ \hline 
Manual Feature-based (Manual) & This baseline only considers the 11 manual features described earlier in Section IV\\ \hline 
Wikipedia Word Embedding (Wiki) & This baselinse only considers the linear classifier trained on the pre-trained Wikipedia Embedding model \\ \hline 
Local Embedding and Manual Feature-based Ensemble (Local-Manual) & This baseline combines \emph{Local} and \emph{Manual} by training two Linear Regression classifiers and weighting their probabilities to get the final result (using the validation set). \\ \hline 
Local Embedding and Wikipedia Word Embedding Ensemble (Wiki-Local) & This baseline combines \emph{Local} and \emph{Wiki} using the same methodology as for \emph{Local-Manual}.\\ \hline 
Wikipedia Word Embedding and Manual Feature-based Ensemble (Wiki-Manual) & This baseline combines \emph{Manual} and \emph{Wiki} using the same methodology as for \emph{Local-Manual}. \\ \hline 
\end{tabular}\label{table:RQ1Nepal}
\end{table*}


\subsubsection{Baselines for Low-supervision Transfer Learning for Urgency Detection}
For RQ2, we consider three baselines besides our own approach:

{\bf Target-only Local (Target Local)}: This baseline is essentially the \emph{Wiki-Manual} baseline described in the previous section and trained on the target dataset (i.e. no transfer learning is used, and no source is assumed). This baseline is used to illustrate the benefits of transfer learning, since this baseline sets the minimum benchmark that has to be bested by a transfer learning baseline. 

{\bf Locally Supervised with Source Embedding (Embedding Transform)}: Similar to our approach on RQ1, manual features, source embeddings and pre-trained Wikipedia embeddings are used to train three classifiers (but on the labeled target domain), and average their probabilities as the final result. While the local embeddings are trained on the source domain (since unlabeled data is not available for the target domain), all classifier training is always done on the target.

{\bf Locally Supervised with Up-sampling and Source Embedding (Upsample)}: This baseline is the same as Embedding Transform, except to boost the power of the baseline, we upsample the labeled data (in the target dataset) by 6x. Thus, this baseline tries to mitigate source bias and concept drift by giving more importance to the transfer domain. This baseline is also more appropriate for the case where the target training data is extremely limited.

\subsection{Results and Discussion}



\begin{table*}
\centering
\caption{Results investigating RQ1 on the Nepal and Kerala datasets.}
\subcaption{Nepal}
\begin{tabular}{|p{1.5in}|p{1.2in}| p{1.2in}| p{1.2in}| p{1.2in}|} \hline
System & Accuracy & Precision & Recall & F-Measure \\ \hline 
Local & $63.97\%$ & $64.27\%$ & $64.50\%$ & $63.93\%$ \\ \hline 
Manual & $64.25\%$ & {\bf 70.84\%}$^{**}$ & $48.50\%$ & $57.11\%$ \\ \hline 
Wiki & $67.25\%$ & $66.51\%$ & $69.50\%$ & $67.76\%$ \\ \hline 
Local-Manual & $65.75\%$ & $67.96\%$ & $59.50\%$ & $62.96\%$ \\ \hline 
Wiki-Local & $67.40\%$ & $65.54\%$ & $68.50\%$ & $66.80\%$ \\ \hline 
Wiki-Manual & $67.75\%$ & $70.38\%$ & $63.00\%$ & $65.79\%$ \\ \hline 
\emph{Our Approach} & {\bf 69.25\%}$^{***}$ & $68.76\%$ & {\bf 70.50\%}$^{**}$ & {\bf 69.44\%}$^{***}$ \\ \hline 
\end{tabular}\label{table:RQ1Nepal}

\bigskip
\subcaption{Kerala}

\begin{tabular}{|p{1.5in}|p{1.2in}| p{1.2in}| p{1.2in}| p{1.2in}|} \hline
System & Accuracy & Precision & Recall & F-Measure \\ \hline 
Local & $56.25\%$ & $37.17\%$ & $55.71\%$ & $44.33\%$ \\ \hline 
Manual & $65.00\%$ & $47.82\%$ & {$\bf 55.77\%$} & $50.63\%$ \\ \hline
Wiki & $63.25\%$ & $42.07\%$ & $46.67\%$ & $44.00\%$ \\ \hline
Local-Manual & $64.50\%$ & $46.90\%$ & $51.86\%$ & $48.47\%$ \\ \hline
Wiki-Manual & $62.25\%$ & $43.56\%$ & $52.63\%$ & $46.93\%$ \\ \hline
Wiki-Manual & {\bf 68.75\%}$^{***}$ & $51.04\%$ & $54.29\%$ & {\bf 52.20\%}$^{**}$ \\ \hline
\emph{Our Approach} & $68.50\%$ & {\bf 51.39\%}$^{***}$ & $52.76\%$ & $51.62\%$ \\ \hline
\end{tabular}\label{table:RQ1Kerala}
\end{table*}



\begin{table*}
\centering
\caption{Results investigating RQ2 using the Nepal dataset as source and Macedonia dataset as target.}
\begin{tabular}{|p{1.5in}|p{1.2in}| p{1.2in}| p{1.2in}| p{1.2in}|} \hline
System & Accuracy & Precision & Recall & F-Measure \\ \hline 
Local & $58.76\%$ & $52.96\%$ & $59.19\%$ & $54.95\%$ \\ \hline 
Transform & $58.62\%$ & $51.40\%$ & {\bf 60.32}$\%^{*}$ & $55.34\%$ \\ \hline 
Upsample & $59.38\%$ & $52.35\%$ & $57.58\%$ & $54.76\%$ \\ \hline 
\emph{Our Approach} & {\bf 61.79\%}$^{*}$ & {\bf 55.08}$\%$ & $59.19\%$ & {\bf 56.90}$\%$ \\ \hline 
\end{tabular}\label{table:RQ2KeralaMacedonia}
\end{table*}

\begin{table*}
\centering
\caption{Results investigating RQ2 using the Kerala dataset as source and Macedonia dataset as target.}
\begin{tabular}{|p{1.5in}|p{1.2in}| p{1.2in}| p{1.2in}| p{1.2in}|} \hline
System & Accuracy & Precision & Recall & F-Measure \\ \hline 
Local & $58.76\%$ & $52.96\%$ & $59.19\%$ & $54.95\%$ \\ \hline 
Transform  & $62.07\%$ & $55.45\%$ & $64.52\%$ & $59.09\%$ \\ \hline
Upsample  & {\bf 64.90}$\%^{***}$ & {\bf 57.98}$\%^{*}$ & {\bf 65.48}$\%^{***}$ & {\bf 61.30}$\%^{***}$ \\ \hline
\emph{Our Approach}  & $62.90\%$ & $56.28\%$ & $62.42\%$ & $58.91\%$ \\ \hline
\end{tabular}\label{table:RQ2?}
\end{table*}

\begin{table*}
\centering
\caption{Results investigating RQ2 using the Nepal dataset as source and Kerala dataset as target.}
\begin{tabular}{|p{1.5in}|p{1.2in}| p{1.2in}| p{1.2in}| p{1.2in}|} \hline
System & Accuracy & Precision & Recall & F-Measure \\ \hline 
Local & $58.65\%$ & {\bf 42.40}$\%$ & $47.47\%$ & $36.88\%$ \\ \hline 
Transform & $53.74\%$ & $32.89\%$ & {\bf 57.47}$\%^{*}$ & $41.42\%$ \\ \hline 
Upsample & $53.88\%$ & $31.71\%$ & $56.32\%$ & $40.32\%$ \\ \hline 
\emph{Our Approach} & {\bf 58.79}$\%$ & $35.26\%$ & $55.89\%$ & {\bf 43.03}$\%^{*}$ \\ \hline 
\end{tabular}\label{table:RQ2?}
\end{table*}

\begin{table*}
\centering
\caption{Results investigating RQ2 using the Kerala dataset as source and Nepal dataset as target.}
\begin{tabular}{|p{1.5in}|p{1.2in}| p{1.2in}| p{1.2in}| p{1.2in}|} \hline
System & Accuracy & Precision & Recall & F-Measure \\ \hline 
Local & $60.26\%$ & {\bf 61.80}$\%$ & $59.94\%$ & $59.88\%$ \\ \hline
Transform & {\bf 61.18}$\%^{*}$ & $61.04\%$ & $63.63\%$ & $62.08\%$ \\ \hline
Upsample & $60.29\%$ & $59.44\%$ & {\bf 66.02}$\%^{*}$ & {\bf 62.50}$\%^{*}$ \\ \hline
\emph{Our Approach} & $60.06\%$ & $59.54\%$ & $63.98\%$ & $61.64\%$ \\ \hline
\end{tabular}\label{table:RQ2?}
\end{table*}


Table IV illustrate the result for RQ1 on the Nepal and Kerala datasets. The results illustrate the viability of urgency detection in low-supervision settings (with our approach yielding 69.44\% F-Measure on Nepal, at 99\% significance compared to the Local baseline), with different feature sets contributing differently to the four metrics. While the local embedding model can reduce precision, for example, it can help the system to improve and accuracy and recall. Similarly, manual features reduce recall, but help the system to improve accuracy and precision (sometimes considerably). To truly address the urgency problem, therefore, a multi-pronged ensemble approach is justified, as also argued intuitively in Section IV. We also note that the pre-trained Wikipedia embedding model proved to be an important tool in improving the generalization ability of the model and not requiring any labeled or unlabeled data; in essence, serving as a free resource that could be helped to regularize and stabilize models that would otherwise be uncertain in low-supervision settings. 

Concerning transfer learning experiments (RQ2), we note that source domain embedding model can improve the performance for target model, and upsampling has a generally positive effect (Tables V-VIII). As expected, transfer learning performance (RQ2) is generally lower compared to the low-supervision urgency detection on a \emph{single} dataset\footnote{The best F-Measure achieved on Nepal in Table IV was more than 69\%, but when using Kerala as source, only 62.5\% F-Measure could be achieved (Table VIII).} (RQ1). Note that at least one of the transfer learning methods always bests the \emph{Local} baseline on all metrics (except precision in Table VII, a result not found to be significant even at the 90\% level). Our approach shows a slight improvement over the upsampling baseline on two of the four scenarios (Tables V and VII) by 2-2.7\% on the F-Measure metric, which shows the diminishing returns from mixing source and target labeled training data.  Further improving performance by high margins will require a radically new approach left for future work.

\section{Conclusion and Future Work}\label{conclusion}

This paper presented minimally supervised urgency detection approaches for short texts (such as tweets) in the aftermath of an arbitrary humanitarian crisis such as the 2015 Nepal earthquake. The presented systems covered two scenarios that often emerge in the real world. In the first scenario, a small amount (a few hundred messages) of training data labeled as urgent or non-urgent is available, along with a copious background corpus. In the second scenario, similar data is available for a `source' domain but not for the target domain (expressing a `new crisis') for which the urgency detection needs to be deployed. As messages are streaming in for this new domain, investigators label a few samples, but cannot rely on the availability of a background corpus since urgency needs to be tagged in real time before the crisis has fully subsided. To accomplish this challenging goal, our approach relies on a simple but robust transfer learning methodology. Experimental results on three real-world datasets validate our methods. 

Some of the obvious avenues for future work are to improve the existing approach incrementally by (for example) adding more manual features and using more sophisticated local embedding model, possibly with more advanced tuning of hyperparameters like the learning rate and vector dimensionality. For improving transfer learning, we are considering using a deep learning model with priors to truly leverage the presence of a source, albeit one covering a domain that is different from the target. Deep learning for transfer learning is still in its infancy in the machine learning community, and has not been demonstrated for difficult and irregular social media datasets. However, we believe that this presents an opportunity for further study.

\section*{Acknowledgements}
The authors gratefully acknowledge the ongoing support and funding of the DARPA
LORELEI program, and our partner collaborators in providing detailed analysis. The
views and conclusions contained herein are those of the authors and should not be interpreted as necessarily
representing the official policies or endorsements, either expressed or implied, of DARPA, AFRL, or the U.S.
Government.

\bibliography{main}
\bibliographystyle{IEEEtran}

\end{document}